\documentclass[letterpaper, 10 pt, conference]{ieeeconf}  % Comment this line out if you need a4paper

\IEEEoverridecommandlockouts                              % This command is only needed if 
\usepackage{epsfig} % for postscript graphics files

\usepackage{graphics} % for pdf, bitmapped graphics files
\usepackage{times} % assumes new font selection scheme installed
\usepackage{amsmath, bm} % assumes amsmath package installed
\usepackage{amsfonts}  % assumes amsmath package installed
\usepackage{color}
\usepackage{cite}

\usepackage{color}
\usepackage[hidelinks]{hyperref} % Uncomment to have clickable cross-ref

\title{\LARGE \bf
Optimal Control for Quadruped Locomotion using LTV MPC
}

\author{Andrew Zheng$^{*1}$, Sriram S.K.S Narayanan$^{*1}$ % <-this % stops a space
\thanks{*These authors contributed equally}% <-this % stops a space
\thanks{$^{1}$Andrew Zheng is a Masters student with the Department of Mechanical Engineering, Clemson University, Clemson, SC 29630, USA
        {\tt\small azheng@clemson.edu}}%
\thanks{$^{1}$Sriram S.K.S Narayanan is a PhD student with the Department of Mechanical Engineering, Clemson University, Clemson, SC 29630, USA
        {\tt\small sriramk@clemsons.edu}}%
}

\begin{document}
\maketitle
\thispagestyle{empty}
\pagestyle{empty}

%%%%%%%%%%%%%%%%%%%%%%%%%%%%%%%%%%%%%%%%%%%%%%%%%%%%%%%%%%%%%%%%%%%%%%%%%%%%%%%%
\begin{abstract}
This paper presents a state-of-the-art optimal controller for quadruped locomotion. The robot dynamics is represented using a single rigid body (SRB) model. A linear time-varying model predictive controller (LTV MPC) is proposed by using linearization schemes. Simulation results show that the LTV MPC can execute various gaits, such as trot and crawl, and is capable of tracking desired reference trajectories even under unknown external disturbances. The LTV MPC is implemented as a quadratic program using $qpOASES$ through the $CasADi$ interface at 50 Hz. The proposed MPC can reach up to 1 m/s top speed with an acceleration of 0.5 m/s$^2$ executing a trot gait. The implementation is available at \url{https://github.com/AndrewZheng-1011/Quad_ConvexMPC}
\end{abstract}

%%%%%%%%%%%%%%%%%%%%%%%%%%%%%%%%%%%%%%%%%%%%%%%%%%%%%%%%%%%%%%%%%%%%%%%%%%%%%%%%
\section{INTRODUCTION}
Quadruped research has seen great success in the research community over the past few years \cite{wensing2022optimization}. This is highlighted through the DARPA Subterranean Challenge, where the top two teams had either ANYmal from ANYbotics or Spot from Boston Dynamics as a core component to their underground exploration challenge \cite{tranzatto2022cerberus}, \cite{chung2022into}. This can be attributed to advancements in both the algorithmic component, researchers finding clever ways to deal with the hybrid dynamics, and the hardware component, hardware having real-time capabilities for nonlinear optimization problems.

Furthermore, there have been great advancements in neural network based models for quadruped locomotion. \cite{feng2022genloco} learns a zero-shot policy for different quadruped models. This is done by learning reliable gaits for different quadruped models during training with the combination of different architectures, such as imitation learning and a standing controller. \cite{agarwal2022legged} uses a vision to desired joint position policy where the authors used a clever scandot approach to train reinforcement policies more efficiently and reliably. And lastly, \cite{miki2022learning} uses a belief encoder and decoder architecture for proprioceptive and exteroceptive sensors to construct a belief state of their environment. The authors then use a student-teacher policy to train this network, with the teacher having access to privileged information and the student using estimated information. However, the general challenges with reinforcement learning based methods are their grey-box nature and lack of stability guarantees.

While the quadruped locomotion community has started to gain traction in various fields of the robotics community, evident with the amount of navigation and reinforcement learning based research using quadrupeds as a mobile base platform \cite{feng2022genloco, agarwal2022legged, miki2022learning, hoeller2021learning, ji2022hierarchical}, optimal control formulations for legged locomotion are still considered state of the art algorithms.

Generally, the control problem in quadruped locomotion consists of designing the required joint torques for the legs to track the desired center of mass reference trajectory. However, due to the hybrid nature of the locomotion problem, there have been numerous approaches that deals with the discontinuous nature of the problem. Traditional methods consisted of using Hybrid Zero Dynamics (HZD) and feedback linearization to ensure stability, and reference tracking capabilities \cite{ames2014rapidly}. However, these control formulations are not very agile and cannot be computed in an online fashion. Furthermore, they are not robust to disturbance and hence cannot typically traverse challenging terrain. 
% However, some of the challenges that are faced when designing optimal controllers for quadruped locomotion are the high dimensionality, nonlinearity, underactuated, and hybrid nature of the dynamical system. Traditional methods within the legged locomotion community consisted of using Hybrid Zero Dynamics (HZD) and feedback linearization to ensure stability and reference tracking capabilities. However, these control formulations are not optimal and do not generate very agile motion. Further, they are not robust to disturbances and hence cannot typically traverse challenging terrain. 

More recent methods propose the use of reduced order simplified models such as single rigid body (SRB) dynamics to model the quadruped’s floating base \cite{winkler18_phd}. This simplifies the original control design problem into a hierarchical control method where the objective of the high-level controller is to find the required ground reaction forces (GRFs) from legs that are instance, typically defined by a gait planner. The simplified model is further linearized to formulate the optimization problem into a quadratic program (QP) which enables real-time computing \cite{di2018dynamic, chignoli2020variational}. The GRFs are then mapped to motor torques using inverse dynamics using low-level control to track the corresponding reference trajectory. As a result, the model predictive control scheme has seen great success in executing motion plans while traversing terrains.

In this project, a state-of-the-art convex model predictive control (MPC) is reformulated for the high-level optimal control design task. A MATLAB simulation platform was used to simulate the locomotion problem based on this work \cite{ding2021representation}. We show the capabilities of our LTV MPC formulation and how the control sequence generated by the optimizer can track our desired reference trajectories.

%%%%%%%%%%%%%%%%%%%%%%%%%%%%%%%%%%%%%%%%%%%%%%%%%%%%%%%%%%%%%%%%%%%%%%%%%%%%%%%%
\section{PRELIMINARIES}
\subsection{Notations}
The cross product of two vectors $\bm{a}$, $\bm{b}$ $\in \mathbb{R}^3$ defined as $\bm{a} \times \bm{b}$ can be represented as a product of skew-symmetric matrix times a vector $\hat{\bm{a}}\bm{b}$. The hat operator $\hat{(\cdot)}: \mathbb{R}^3 \rightarrow{} \bm{\mathfrak{so}}(3)$ maps the elements of the vector $\bm{a} = [a_1 \hspace{1mm} a_2 \hspace{1mm} a_3]^\top$ to $\bm{\mathfrak{so}}(3)$, the lie algebra of $\bm{SO}$(3) at identity as 

\begin{equation}
\hat{\bm{a}} = 
    \begin{bmatrix}
    0 &-a_3 &a_2 \\
    a_3 &0 &-a_1 \\
    -a_2 &a_1 &0 
    \end{bmatrix}
    \in \bm{\mathfrak{so}}(3)
\end{equation}

% The inverse hat operator $(\cdot)^\vee: \mathfrak{so}(3) \xrightarrow{} \mathbb{R}^3$ maps an element of $\mathbf{\mathfrak{so}}$(3) back to a vector in $\mathbb{R}^3$ such that $\hat{\bm{a}}^\vee = \bm{a}$. The lie algebra of $\mathfrak{so}(3)$ is a vector space whose elements can be decomposed as 
% \begin{align}
%     &\hat{\bm{a}} = a_1 \bm{E_x} + a_2 \bm{E_y} + a_3 \bm{E_z} \\
%     &\bm{E_x} = \begin{bmatrix} 0 &0 &0 \\ 0 &0 &-1 \\ 0 &1 &0 \end{bmatrix} \nonumber \\
%     &\bm{E_y} = \begin{bmatrix} 0 &0 &1 \\ 0 &0 &0 \\ -1 &0 &0 \end{bmatrix} \nonumber \\
%     &\bm{E_z} = \begin{bmatrix} 0 &-1 &0 \\ 1 &0 &0 \\ 0 &0 &0 \end{bmatrix} \nonumber
%     \end{align}
% Let, $skew_n$ denote the space of $n \times n$ skew-symmetric matrices, then $Skew(\cdot)$ extracts the skew-symmetric part of a $n \times n$ matrix.
% \begin{equation}
%     skew(A) = \frac{1}{2}(A-A^\top)
% \end{equation}

\subsection{Rigid Body Dynamics}
\begin{figure}
    \centering
    \includegraphics[width=1\linewidth]{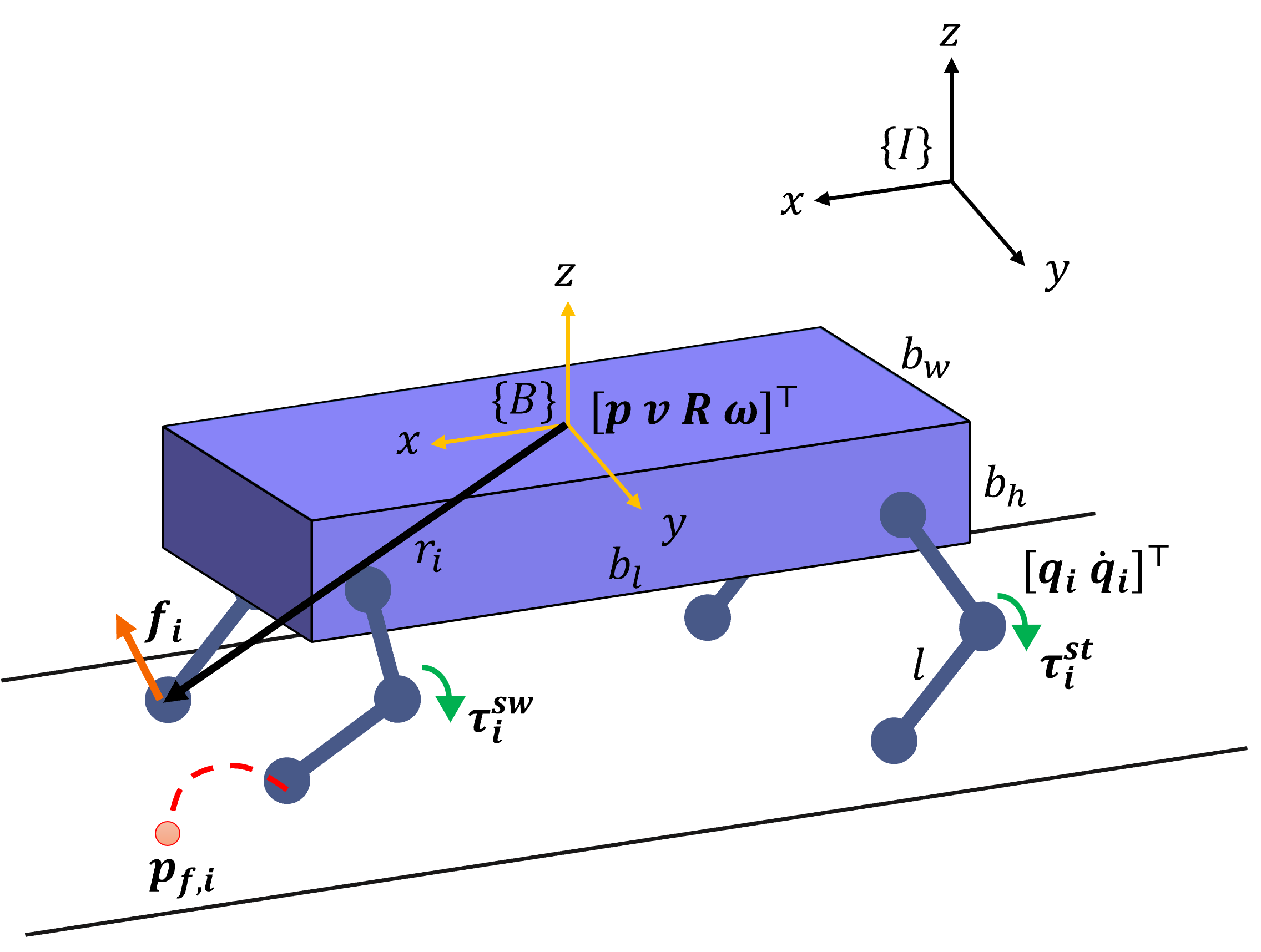}
    \caption{Quadruped locomotion description}
    \label{fig:quadl}
\end{figure}

Quadruped locomotion can be described as a hybrid system switching between swing and stance phase dynamics. The switching logic is determined by a contact detection algorithm. This system is under-actuated since there is no direct actuation along the direction of motion. The robot must exert ground reaction forces (GRFs) at each foot in contact to propel its base forward to follow a reference trajectory. The rigid body model defines the evolution of the states of the center of mass due to the applied GRFs, as shown in Figure \ref{fig:quadl}. The state vector is defined by 

\begin{equation} \label{eq:mapping}
    \bm{x} := [\bm{p} \hspace{1mm} \bm{v} \hspace{1mm} \bm{\mathbf{R}} \hspace{1mm} \bm{\omega}]^\top \in \mathbb{R}^{18}   
\end{equation}
where $\bm{p} \in \mathbb{R}^3$ is the Cartesian position and $\bm{v} \in \mathbb{R}^3$ is the velocity of the robot's center of mass,  $\bm{R} \in \bm{SO}(3)$ is the $3 \times 3$ rotation matrix and $\bm{\omega} \in \mathbb{R}^3$ is the angular velocity. Each leg $i$ of the robot generates the GRFs $\bm{f_i} \in \mathbb{R}^3$ which leads to a net external wrench $\bm{\mathcal{F}} \in \mathbb{R}^6$ given by

\begin{equation}
    \bm{\mathcal{F}} = \begin{bmatrix} \bm{F} \\ \bm{\tau} \end{bmatrix} = \begin{bmatrix} \mathbb{I} & \mathbb{I} & \mathbb{I} & \mathbb{I} \\ \hat{\bm{r}}_1 & \hat{\bm{r}}_2 & \hat{\bm{r}}_3 & \hat{\bm{r}}_4 \end{bmatrix} \begin{bmatrix} \bm{f}_1 \\ \bm{f}_2 \\ \bm{f}_3 \\ \bm{f}_4 \end{bmatrix}
\end{equation}

Here, $\bm{\mathcal{F}} \in \mathbb{R}^3$ is the net force acting on the rigid body, $\bm{\tau} \in \mathbb{R}^3$  is the net torque in the inertial frame $\mathcal{I}$ with $\mathbf{r_i}$ denoting the $i$th foot location relative to the center of mass in inertial frame $\mathcal{I}$ and $\mathbb{I}$ is a $3 \times 3$ identity matrix. The SRB dynamics are given by 
\begin{subequations}
\begin{align}
    \dot{\bm{p}} &= \bm{v}  \label{eq:lin1}\\
    \dot{\bm{v}} &= \frac{1}{m} \bm{F} - \bm{g}  \label{eq:lin2}\\
    \dot{\bm{R}} &= \bm{R} \cdot ^{\mathcal{B}}\hat{\bm{\omega}}  \label{eq:ang1}\\
    ^{\mathcal{B}}\dot{\bm{\omega}} &= ^{\mathcal{B}}\bm{I}^{-1}\left(^{\mathcal{B}} \bm{\tau} - ^{\mathcal{B}}\bm{\omega} \times ^{\mathcal{B}}\bm{I} ^{\mathcal{B}}\bm{\omega} \right) \label{eq:ang2}
    %&\dot{\bm{\omega}} = \bm{I}^{-1}\left(\bm{R}^T\left(\sum_{i=1}^n \bm{r_i} \times \bm{f_i}\right) \right) \label{ang2}%
\end{align}
\end{subequations}
where $m$ is the mass of the robot, $^{\mathcal{B}} \bm{I} \in \bm{R}^{3 \times 3}$ moment of inertia of the robot, $\bm{g} = [0 \hspace{2mm} 0 \hspace{2mm} g]^\top$ denotes the gravity vector Here $^{\mathcal{B}} \bm{\omega}$ and $^{\mathcal{B}} \bm{I}$ are defined in the body frame $\mathcal{B}$ while $\bm{p}$, $\bm{v}$, $\bm{f_i}$, $\bm{R}$ are defined in the inertial frame $\mathcal{I}$.

\subsection{Robot Leg Dynamics}
Consider a quadruped robot with $m$ joints in each foot $i$. Then a joint state and joint velocity vectors can be defined as
\begin{subequations}
\begin{align}
    &\bm{q}_i:=[q_1 \hspace{1mm} q_2 \hspace{1mm} \hdots \hspace{1mm} q_m]_i^\top \in \mathbb{R}^m \\
    &\dot{\bm{q}}_i:=[\dot{q}_1 \hspace{1mm} \dot{q}_2 \hspace{1mm} \hdots \hspace{1mm} \dot{q}_m]_i^\top \in \mathbb{R}^m
\end{align}
\end{subequations}
During the swing phase of robot locomotion, when the robot leg is not in contact with the ground, the dynamics can be defined based on the Lagrangian formulation as
\begin{equation}
    \bm{M_i}(\bm{q}_i)\ddot{\bm{q}_i} + \bm{h_i}(\bm{q}_i,\dot{\bm{q}}_i) = \bm{\tau}_i^{sw}
\end{equation}
Here $\bm{q}_i$, $\dot{\bm{q}}_i$ and $\ddot{\bm{q}}_i$ are the joint state, velocity, and acceleration, respectively, $\bm{M}_i(\bm{q}_i) \in \mathbb{R}^m$ is the mass matrix of the leg, $\bm{h}_i(\bm{q}_i,\dot{\bm{q}}_i) \in \mathbb{R}^m$ represents the Coriolis and gravity terms acting on the leg, and $\bm{\tau}_i^{sw} \in \mathbb{R}^m$ is a feed-forward joint torque input.

%%%%%%%%%%%%%%%%%%%%%%%%%%%%%%%%%%%%%%%%%%%%%%%%%%%%%%%%%%%%%%%%%%%%%%%%%%%%%%%%
\section{Linear Time-Varying Model Predictive Control (LTV MPC)}
\subsection{Linear Time-Varying Dynmaics} \label{subsection: LTV Dynamics}
The orientation of the robot can be represented using Euler angles $\bm{\Theta} = [\phi, \theta, \psi]$ where $\phi$ is the pitch angle about the x-axis, $\theta$ is the roll angle about the y-axis, and $\psi$ is the yaw angle about the z-axis. To represent the body frame $\mathcal{B}$ in the inertial frame $\mathcal{I}$, a transformation matrix can be constructed using these angles as follows
\begin{equation}
    \bm{R} = \bm{R}_z(\psi)\bm{R}_y(\theta)\bm{R}_x(\phi)
\end{equation}
where each $\bm{R}_i(\alpha)$ represented the rotation of angle $\alpha$ about axis $i$. In this work, we assume that the robot is walking in a straight line without minimal body roll and pitch (i.e., $\theta, \phi \approx 0$). Further, we use the small angle assumption to rewrite \eqref{eq:ang1} in a linear form as
\begin{align}
    \begin{bmatrix}\dot{\phi} \\ \dot{\theta} \\ \dot{\psi}\end{bmatrix} &\approx \bm{R}_z(\psi)\bm{w} \\
    \bm{R}_z(\psi) &= \begin{bmatrix}
        \cos\psi & -\sin\psi & 0 \\
        \sin\psi & \cos\psi & 0 \\
        0 & 0 & 1
    \end{bmatrix} \nonumber
\end{align}
For a rigid body with small angular velocities, $\bm{\omega} \times \bm{I} \bm{\omega}$ term can be neglected in \eqref{eq:ang2}. This assumption discards the effects of angular motion, such as precession and mutation, and has been used to develop controllers for quadruped robots such as \cite{di2018dynamic} and \cite{focchi2017high}. Hence \eqref{eq:ang2} can be linearized as
\begin{equation}
    ^{\mathcal{B}}\dot{\bm{\omega}} = ^{\mathcal{B}}\bm{I}^{-1} \hspace{1mm}^{\mathcal{B}} \bm{\tau}
\end{equation}
Note that $^{\mathcal{B}}\bm{w}$, $^{\mathcal{B}}\bm{I}$ and $^{\mathcal{B}}\bm{\tau}$ are defined in the body frame. For the convenience of computations, these can be converted to the world frame as follows
\begin{subequations}
\begin{align}
    \bm{I} &\approx \bm{R}_z(\psi) ^{\mathcal{B}} \bm{I} \bm{R}_z(\psi)^\top \\
    \bm{w} &\approx \bm{R}_z(\psi) ^{\mathcal{B}} \bm{w} \\
    \bm{\tau} &\approx \bm{R}_z(\psi) ^{\mathcal{B}} \bm{\tau}
\end{align}
\end{subequations}
The simplified linear time-varying dynamics can be written as
\begin{align}
    \begin{bmatrix}
        \dot{\bm{p}} \\ \dot{\bm{v}} \\ \dot{\bm{\Theta}} \\ \dot{\bm{w}}
    \end{bmatrix} & \approx 
    \begin{bmatrix}
        \bm{0}_3 & \bm{0}_3 & \bm{0}_3 & \bm{1}_3 \\
        \bm{0}_3 & \bm{0}_3 & \bm{0}_3 & \bm{0}_3 \\
        \bm{0}_3 & \bm{0}_3 & \bm{R}_z(\psi) & \bm{0}_3 \\
        \bm{0}_3 & \bm{0}_3 & \bm{0}_3 & \bm{0}_3
    \end{bmatrix} \begin{bmatrix}
        \bm{p} \\ \bm{v} \\ \bm{\Theta} \\ \bm{w}
    \end{bmatrix} +  \nonumber \\
    &\begin{bmatrix}
        \bm{0}_3 & \hdots{} & \bm{0}_3 \\
        \bm{1}_3/m & \hdots{} & \bm{1}_3/m \\
        \bm{0}_3 & \hdots{} & \bm{0}_3 \\
        \bm{I}^{-1}\hat{\bm{r}}_1 & \hdots{} &  \bm{I}^{-1}\hat{\bm{r}}_n
    \end{bmatrix} \begin{bmatrix}
        \bm{f}_1 \\ \vdots{} \\ \bm{f}_n
    \end{bmatrix} + \begin{bmatrix}
        \bm{0}_3 \\ \bm{0}_3 \\ \bm{0}_3 \\ \bm{g}
    \end{bmatrix}
\end{align}
The above equation can be converted to a standard time-varying state-space form by embedding gravity $\bm{g}$ as a state to get
\begin{align} \label{eqn:c_LTV}
    &\dot{\bm{x}}(t) = \bm{A}_c(\psi)\bm{x}(t) + \bm{B}_c(\bm{r}_1,\hdots,\bm{r}_n, \psi)\bm{u}(t) \\
    &\bm{A}_c(\psi) = \begin{bmatrix}
        \bm{0}_3 & \bm{0}_3 & \bm{0}_3 & \bm{1}_3 & \bm{0}_3  \\
        \bm{0}_3 & \bm{0}_3 & \bm{0}_3 & \bm{0}_3 & [0 \hspace{2mm} 0 \hspace{2mm} 1]^\top \\
        \bm{0}_3 & \bm{0}_3 & \bm{R}_z(\psi) & \bm{0}_3 & \bm{0}_3\\
        \bm{0}_3 & \bm{0}_3 & \bm{0}_3 & \bm{0}_3 & \bm{0}_3 \\
        \bm{0}_3 & \bm{0}_3 & \bm{0}_3 & \bm{0}_3 & \bm{0}_3
        \end{bmatrix} \nonumber \\
    &\bm{B}_c(\bm{r}_1,\hdots,\bm{r}_n, \psi) = \begin{bmatrix}
        \bm{0}_3 & \hdots{} & \bm{0}_3 & \bm{0}_3 \\
        \bm{1}_3/m & \hdots{} & \bm{1}_3/m & \bm{0}_3\\
        \bm{0}_3 & \hdots{} & \bm{0}_3 & \bm{0}_3\\
        \bm{I}^{-1}\hat{\bm{r}}_1 & \hdots{} &  \bm{I}^{-1}\hat{\bm{r}}_n & \bm{0}_3\\
         \bm{0}_3 & \hdots{} & \bm{0}_3 & \bm{0}_3
        \end{bmatrix} \nonumber
\end{align}
Here, $\bm{x} = \begin{bmatrix} \bm{p} & \bm{v} & \bm{\Theta} & \bm{w} & \bm{g} \end{bmatrix}^\top \in \mathbb{R}^{13}$, $\bm{A}_c(\psi) \in \mathbb{R}^{13 \times 13}$ is a time-varying function of $\psi$, the yaw angle along the robot trajectory and $\bm{B}_c(\bm{r}_1,\hdots,\bm{r}_n, \psi) \in \mathbb{R}^{13 \times 3n}$ is a time-varying function of the number of feet in contact $\bm{r}_n$ and the yaw angle $\psi$. Further, the size of $\bm{B}_c$ changes depending on $n$ feet in contact, and hence it is more memory efficient to formulate a time-varying MPC with $n$ foot forces as control inputs $\bm{u}(t)$ using the dynamics presented in equation \eqref{eqn:c_LTV}.

\subsection{LTV MPC Formulation}
The mapping of foot forces to the net force and torques acting on the body (given by equation \eqref{eq:mapping}) is not unique. Further, the single rigid body dynamics is underactuated with six degrees of freedom ($\bm{p}$, $\bm{\Theta}$) but only $n\leq 4$ control inputs ($\bm{f}_i$ for $n$ feet in contact). Hence designing an optimal controller based on traditional methods is a challenging problem. In this work, we design an optimal controller based on an MPC approach. The role of the MPC is to design individual foot forces for $n$ feet in contact, which propel the rigid body to follow a given reference trajectory. This acts as a low-level plan which is then converted to joint torques as input for the low-level controller presented in section \ref{stance}. 
In general, an MPC problem with a horizon $N$ can be formulated as

\begin{subequations}
\begin{align}
    \min_{\bm{x},\bm{u}} \hspace{2mm} &\sum_{i=0}^{N-1} \bigl\{||\bm{x}_{k+1} - \bm{x}_{k+1,ref}||_{\bm{Q}_k} + ||\bm{u}||_{\bm{K}_k} \bigr\} \label{eq:opt} \\
    \textrm{subject to}  \nonumber \\
    &\bm{x}_{k+1} = \bm{A}_k \bm{x}_k + \bm{B}_k \bm{u}_k  \label{eq:dynamics} \\
    & \underline{\bm{c}}_k \leq \bm{C}_k \bm{u}_k \leq \bar{\bm{c}}_k \label{eq:constraint1} \\ 
    & \bm{D}_k \bm{u}_k = 0 \label{eq:constraint2}
\end{align}
\end{subequations}

where $\bm{x}_k$ and $\bm{u}_k$ are the state and control input respectively at time step $k$, $\bm{Q}_k$ and $\bm{K}_k$ are diagonal positive definite weight matrices, $\bm{A}_k$ and $\bm{B}_k$ represent the discrete-time system dynamics given in equation \eqref{eq:dynamics}. The constraints on the foot forces are represented using equation \eqref{eq:constraint1}, and \eqref{eq:constraint2}, where $\bm{C}_k$ is the constraint matrix for the feet in contact, \underbar{$\bm{c}$}$_k$  and $\bar{\bm{c}}_k$ are the upper and lower bounds on the corresponding control inputs and $\bm{D}_k$ selects the feet in swing and sets the foot force to zero for the corresponding feet.

\subsection{Constraints}
For the quadruped locomotion problem, the inequality constraint in equation \eqref{eq:constraint1} represents the friction constraint and bounds on the z-direction force exerted by the robot. In general, the friction constraint is described using a friction cone as follows,
\begin{equation}
    \sqrt{\bm{f}_{x,i}^2 + \bm{f}_{y,i}^2} \leq \mu \bm{f}_{z,i}
\end{equation}
where $\mu$ is the coefficient of friction for $i$th feet in contact. This constraint is nonlinear and hence is not admissible for a QP formulation. Hence we use a friction pyramid \cite{trinkle1997dynamic}, which is a linearized version of this constraint. Hence the constraint equation in \eqref{eq:constraint1} reduces to
\begin{subequations}
\begin{align}
    f_{min} &\leq \bm{f}_{z,i} \leq f_{max} \\
    -\mu \bm{f}_{z,i} &\leq \pm \bm{f}_{x,i} \leq \mu \bm{f}_{z,i}\\
    -\mu \bm{f}_{z,i} &\leq \pm \bm{f}_{y,i} \leq \mu \bm{f}_{z,i}
\end{align}   
\end{subequations}
Writing this in standard form $\bm{A}_{ineq,i} \bm{u}_i \leq \bm{b}_{ineq,i}$ for the $i$th feet in contact as

\begin{equation}
    \begin{bmatrix}
        -1 & 0  & -\mu \\
         1 & 0  & -\mu \\
         0 & -1 & -\mu \\
         0 & 1  & -\mu \\
         0 & 0  & -1 \\
         0 & 0  &  1 \\
    \end{bmatrix} \bm{u}_i =
    \begin{bmatrix}
       0 \\ 0 \\ 0 \\  0 \\ -\bm{f}_{z,i,lb} \\ -\bm{f}_{z,i,ub}
   \end{bmatrix}
\end{equation}

where the control bounds \underbar{$\bm{c}$}$_i$ and $\bar{\bm{c}}_i$ are replaced by the 
lower bound $\bm{f}_{z,i,lb}$ and upper bound $\bm{f}_{z,i,ub}$ respectively. For $n$ feet in contact, $\bm{A}_{ineq,n}$ can be formed as a block diagonal matrix with $n$ block for $\bm{A}_{ineq,i}$. Similarly, $\bm{b}_{ineq,n}$ can be formed using $n$ repetitions of $\bm{b}_{ineq,i}$. Further, the equality constraint in equation \eqref{eq:constraint2} can be handled implicitly by formulating the dynamics in \eqref{eq:dynamics} with a time-varying control matrix $\bm{B}_i$ as described in \eqref{eqn:c_LTV}. This reduces the size of the MPC problem and hence is more memory efficient for transferring to the embedded controllers on a physical robot.

\subsection{Reference Trajectory Generation}
Reference trajectories play an important role in high dimensional optimization problems as it guides the search space. By generating reasonable trajectories from simplified models, the optimizer then finds a control law that best follows the trajectories while following constraints. 

To generate reference trajectories, a desired final state was defined. Then, reference trajectories were generated using a kinematic model under zero-order hold assumption until the trajectories generated met the desired final state. 
\begin{subequations}
    \begin{align}
            \bm{p}_{ref} &= \begin{cases}
            \frac{1}{2} \bm{\dot{v}}_d t^2     & 0 \leq t \leq \frac{\bm{v}_d}{\bm{\dot{v}}_d} \\
            \bm{v}_d t - \frac{1}{2}\bm{v}_d\frac{\bm{v}_d}{\bm{a}_d} & \textrm{otherwise}
            \end{cases}\\
           \bm{v}_{ref} &= \begin{cases} 
            \bm{\dot{v}}_d t      & 0 \leq t \leq \frac{\bm{v}_d}{\bm{\dot{v}}_d} \\
            \bm{v}_{ref}    & \textrm{otherwise}
            \end{cases} \\
            \psi_{ref} &= \begin{cases} 
            \omega_{\psi,d} t      & \psi \leq \psi_d \\
            \psi_d    & \textrm{otherwise}
            \end{cases} \\
            \omega_{\psi,ref} &= \begin{cases} 
            \dot{\omega}_{\psi,d} t      & \psi \leq \psi_d \\
            0    & \textrm{otherwise}
            \end{cases}
    \end{align}
\end{subequations}
where $\bm{v}_d$, $\bm{\dot{v}}_d$ are the desired velocity and acceleration for the robot and $\psi_d$, $\omega_{\psi,d}$, $\dot{\omega}_{\psi,d}$ are the desired yaw angle, yaw velocity and acceleration respectively.

\subsection{QP formulation}

%% Figure for hierarchical control section
\begin{figure*}
  \includegraphics[width=\textwidth]{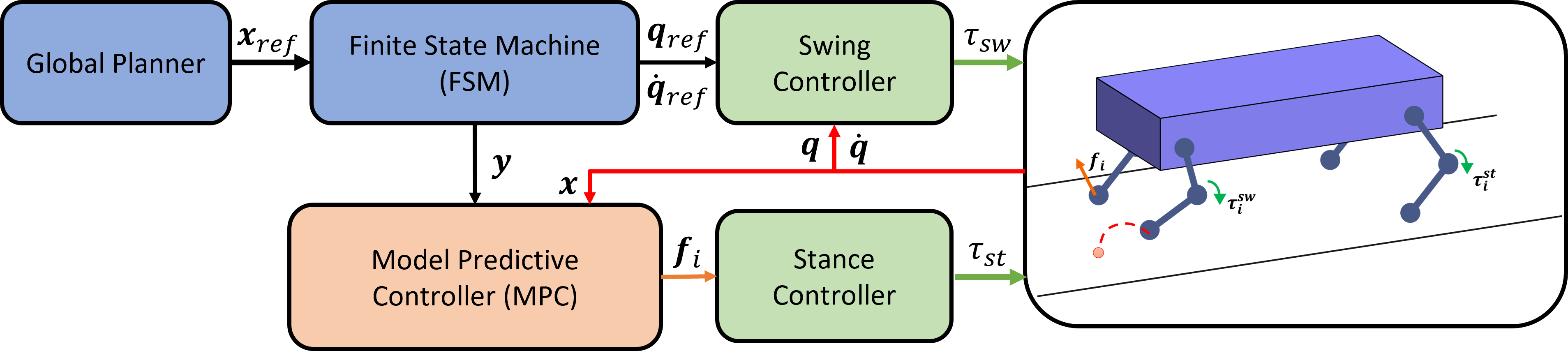}
  \caption{Hierarchical control structure for quadruped control}
\end{figure*}

Since the optimization problem presented in \eqref{eq:opt} has a quadratic cost with linear constraints, the solution can be obtained using a quadratic program (QP) with only $\bm{u}_i$ as the control variable (single shooting). For locomotion with minimal change in roll and pitch angle (which is the case for locomotion in flat terrain), it is reasonable to assume that the linearization scheme will be valid for short horizon lengths. Hence the dynamics constraint in \eqref{eq:dynamics} can be integrated with the cost function as follows
\begin{subequations}
\begin{align}
    J(\bm{U}) &= ||\bm{X} - \bm{X}_{ref}||_{\bar{\bm{Q}}} + ||\bm{U}||_{\bar{\bm{K}}} \\
    \bm{X} &= \bm{A}_{qp}\bm{x}_0 + \bm{B}_{qp}\bm{U} \\
    \bm{A}_{qp} &= \begin{bmatrix}
        \bm{A} & \bm{A}^2 & \hdots{} & \bm{A}^k 
    \end{bmatrix} ^\top \nonumber \\
    \bm{B}_{qp} &= \begin{bmatrix}
        \bm{B} & \bm{0} &\hdots &\bm{0} \\
        \bm{A}\bm{B} & \bm{B} & \hdots &\bm{0}\\
        \vdots{} & \vdots{} & \ddots & \vdots{} \\
        \bm{A}^{k-1}\bm{B} & \bm{A}^{k-2}\bm{B} & \hdots &\bm{B} \nonumber
    \end{bmatrix}
\end{align}
\end{subequations}
where $\bar{\bm{Q}} \in \mathbb{R}^{13k \times 13k}$ is a block diagonal matrix of weights for state deviations $\bm{Q}_k$, $\bar{\bm{K}} \in \mathbb{R}^{3nk \times 3nk}$ is a block diagonal matrix of weights for control magnitude $\bm{K}_k$, $\bm{X}$ and $\bm{U}$ are stacked vectors of state and control inputs $\bm{x}_i$ and $\bm{u}_i$ respectively over the prediction horizon $N$. $\bm{A}_{qp} \in \mathbb{R}^{13N \times 13}$ and $\bm{B}_{qp} \in \mathbb{R}^{13N \times 3nN}$ are the stacked matrices for state and control inputs. A zero-order hold is applied on the matrices $\bm{A}_c$ and $\bm{B}_c$ in \eqref{eqn:c_LTV} to obtain the discretized $\bm{A}$ and $\bm{B}$ matrices respectively. Hence the optimization problem in \eqref{eq:opt} can be written in a standard QP form as follows
\begin{subequations}
\begin{align}
    \min_{\bm{U}} \hspace{5mm} &\frac{1}{2}\bm{U}^\top \bm{H} \bm{U} + \bm{U}^\top \bm{G} \\
    \textrm{subject to} \hspace{5mm} & \bm{A}_{ineq} \bm{U} \leq \bm{b}_{ineq} 
\end{align}
\end{subequations}
where $\bm{A}_{ineq}$ and $\bm{b}_{ineq,e}$ are stacked block diagonal matrices of constraints $\bm{A}_{ineq,n}$ and $\bm{b}_{ineq,n}$ respectively for $n$ legs in contact, $\bm{H} \in \mathbb{R}^{3nN \times 3nN}$ and $\bm{G} \in \mathbb{R}^{3nN \times 3nN}$ are the QP matrices given by
\begin{subequations}
    \begin{align}
        &\bm{H} = 2(\bm{B}_{qp}^\top \bm{\bar{Q}} \bm{B}_{qp} + \bar{\bm{K}}) \\
        &\bm{G} = 2\bm{B}_{qp}^\top \bm{\bar{Q}} (\bm{A}_{qp}\bm{x}_0 - \bm{y})
    \end{align}
\end{subequations}
and $\bm{y}$ is a stacked vector for $\bm{x}_{ref}$ for horizon length $k$.

%%%%%%%%%%%%%%%%%%%%%%%%%%%%%%%%%%%%%%%%%%%%%%%%%%%%%%%%%%%%%%%%%%%%%%%%%%%%%%%%
\section{HIERARCHICAL CONTROL}
\subsection{Swing Leg Control (Inverse Dynamics)}
The control input for swing legs, $\bm{\tau}_i^{sw} \in \mathbb{R}^3$ for each leg typically uses a combination of a feed-forward torque $\bm{\tau}_i^{ff,sw}$ (using inverse dynamics) and PD control to follow the desired leg trajectory. The next touchdown location of each swing foot is given by the Raibert heuristic. This well-known heuristic ensures that each foot lands below its corresponding hip at the middle of the stance phase with a stance duration of $t_{st}$ assuming the robot moves at constant velocity $\bm{v}_{cur}$. When the robot needs to accelerate to follow the desired reference velocity $\bm{v}_{ref}$ (and angular velocity $\bm{\omega}_{cmd}$), the heuristic adjusts the next footstep prediction $\bm{p}_{f,i}$ using feedback terms $k_{raibert}$ and $k_{centrifugal}$
\begin{align}
    \bm{p_{f,i}} = & \bm{p}_{i, hip} + 0.5t_{st}\bm{v}_{cur} + k_{raibert}(\bm{v}_{cur}-\bm{v}_{ref})
\end{align}
where $\bm{p}_{i, hip}$ is the position of the $i$th hip in world coordinates. The gains can be chosen as $k_{raibert}=\sqrt{h/g}$, where $h$ is the height of the base and $g$ is gravity.

A reference trajectory in Cartesian space is generated by interpolating the current footstep location and the predicted touchdown location $\bm{r_{i}}$ (using cubic splines or Hermite polynomials). Using this, the reference trajectory for joint positions $\bm{q}_{ref}$ and velocities $\dot{\bm{q}}_{ref}$ for each joint of the swing leg $i$ can be computed using inverse kinematics. Note that these reference trajectories are in the body reference frame $\mathcal{B}$.
\begin{align} \label{eq:swing}
    &\bm{\tau}_i^{sw} = \bm{M}_i(\bm{q}_i)\ddot{\bm{q}}_i + \bm{h}_i(\bm{q}_i,\dot{\bm{q}}_i) \nonumber \\
    &+ \bm{K}_p^{sw}(\bm{q}_{i, ref}-\bm{q}_i) + \bm{K}_d^{sw}(\dot{\bm{q}}_{i, ref}-\dot{\bm{q}}_i)
\end{align}
The control objective is to track a desired reference trajectory for each swing foot and land at the next touchdown location. Note that there is no change in the position/orientation of the center of mass $\bm{x}(t)$ when executing this controller. It does not require information about the $\bm{x}(t)$ and only uses the joint state values $\bm{q}_i$ and $\dot{\bm{q}}_i$ of each leg. Similarly, the stance phase joint torques $\bm{\tau}_i^{st}$ are computed using \eqref{tau_stance} detailed in  the next section.

\subsection{Stance Leg Control (MPC)} \label{stance}
The objective of the stance phase controller is to generate the required GRFs to propel the base forward to track a given base reference trajectory. The $\bm{f}_i$ (GRFs) output from MPC controller is converted to joint torques for using the relation
\begin{equation}
    \bm{\tau}_i^{st} = \bm{J}_i^\top \bm{R}_i^\top \bm{f}_i \label{tau_stance}
\end{equation}
where $\bm{\tau}_i^{st} \in \mathbb{R}^3$ is the joint torques for stance phase,  $\bm{J}_i \in \mathbb{R}^{3 \times 3}$ is the leg jacobian for the $i$th feet in stance and $\bm{R}_i$ is the corresponding rotation matrix.

\subsection{Finite State Machine (FSM)}
To incorporate a contact-dependent or time-varying MPC, and correspondingly integrate either a tracking controller to track a reference leg trajectory or a stance controller to generate desired GRFs, a finite state machine is constructed.

FSM is a gait schedule planner that is able to schedule gait sequences to each leg, given an adjustment in schedule timings. More formally, given a schedule, a leg phase independent variable is defined to be either in the swing or stance phase. This is defined as $s_i := \{\bar{t}/t^j | j \in \{st, sw\}\}$ where $\bar{t}$ is the dwell time. Transitions between the swing and stance states are defined by the Guard set $G_i:=\{\bar{t} | \bar{t} = T^j\}$ and the reset map $\Delta_j(\bar{t}) = 0$, which resets $s_i$ and $\bar{t}$. Furthermore, the FSM allows different gait motions, such as trotting and crawling. The gait parameters used in this simulation are listed in Table \ref{tab:gait}

\begin{table}[ht]
\caption{Gait parameters}
\centering
\begin{tabular}{ c c c  } 
 \hline
 Gait & Stance time (s) & Swing time (s)\\
\hline
 trot       & 0.1       & 0.18 \\
 bound      & 0.12      & 0.12 \\
 pacing     & 0.08      & 0.2 \\
 gallop     & 0.08      & 0.2 \\
 trot run   & 0.12      & 0.2 \\
 crawl      & 0.3       & 0.1 \\
 \hline
\end{tabular}
\label{tab:gait}
\end{table}

%%%%%%%%%%%%%%%%%%%%%%%%%%%%%%%%%%%%%%%%%%%%%%%%%%%%%%%%%%%%%%%%%%%%%%%%%%%%%%%%
\section{Results and Discussion}
The robot parameters used for simulation are listed in Table \ref{tab:params}. The proposed MPC is implemented in simulation using MATLAB's QP solver quadprog and $qpOASES$ implemented using $CasADi$. The horizon length $N$ was around one gait cycle  with a prediction time step of $0.02$ seconds, i.e., predictions were computed at the rate of 50 Hz. For a trotting gait with stance time $t_{st} = 0.1$ seconds and swing time $t_{sw} = 0.18$ seconds, the horizon length $N$ was set to 15 to match the gain cycle duration of $0.28$ s. The low-level controller typically runs at a much higher frequency (around 1000 Hz). However, it is not part of the simulation results presented in this section. The weights for the MPC cost function in \eqref{eq:opt} and the gains for the swing controller in \eqref{eq:swing} are listed in Table \ref{tab:gains}

\begin{figure}
    \centering
    \includegraphics[width=\linewidth]{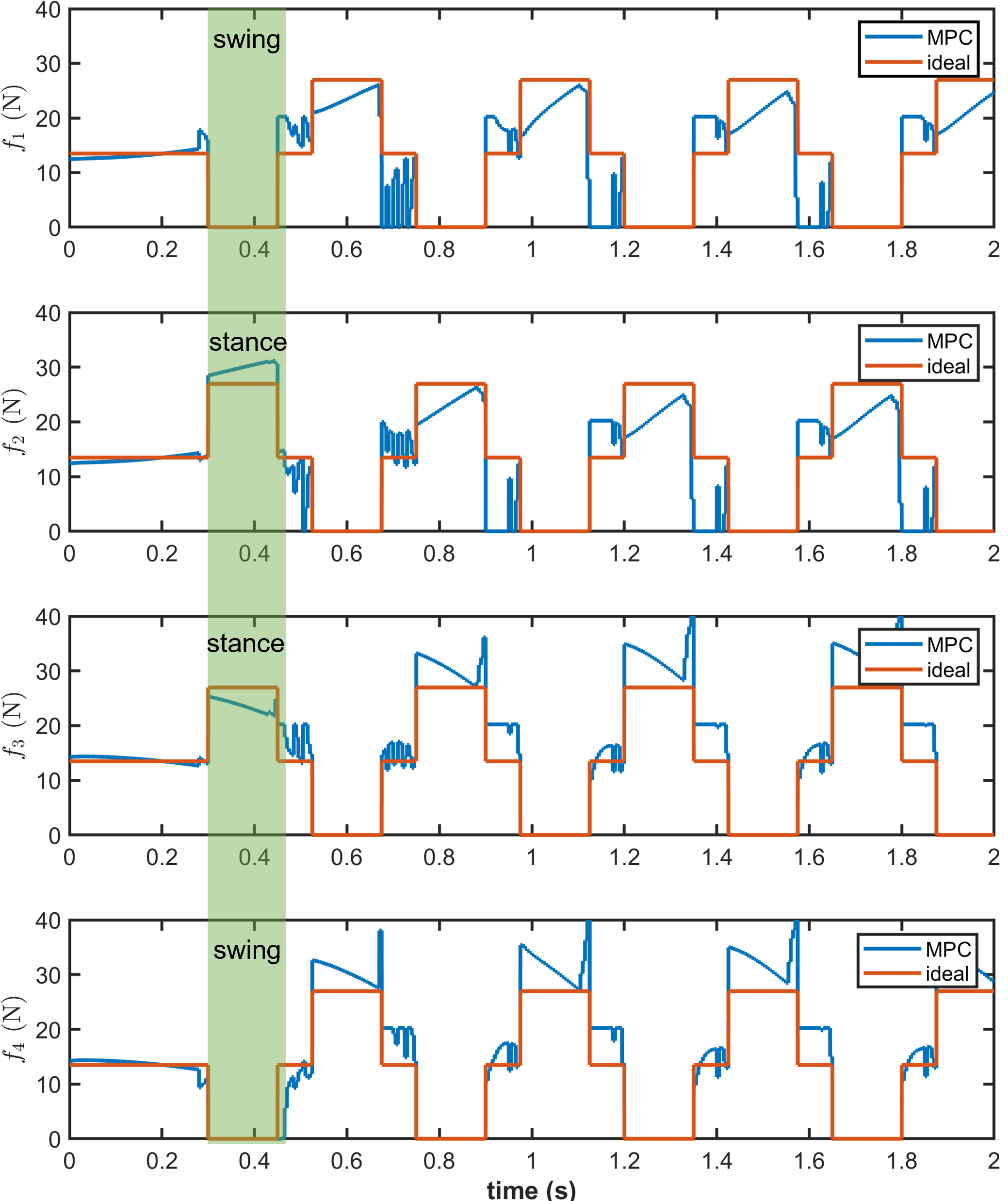}
    \caption{Z-direction GRFs for each leg while executing a trot gait for robot walking in a straight line.}
    \label{fig:trot_z}
\end{figure}

\begin{table}[ht]
\caption{Robot parameters}
\centering
\begin{tabular}{ c c c c } 
 \hline
 Parameter & Variable & Value & Units \\
\hline
 mass & $m$ & 5.5 & kg \\ 
 inertia        & $I_{xx}$  & 0.026 & kg-m$^2$ \\ 
 inertia        & $I_{yy}$  & 0.112 & kg-m$^2$ \\ 
 inertia        & $I_{zz}$  & 0.075 & kg-m$^2$ \\
 body length    & $b_l$     & 0.3   & m \\
 body width     & $b_w$     & 0.088 & m \\
 body height    & $b_h$     & 0.05  & m \\
 link length    & $l$       & 0.14  & m \\    
 nominal height & $z_0$     & 0.2   & m \\
 gravity        & $g$       & 9.81  & m/s$^2$ \\
 friction       & $\mu$     & 1     & [-] \\  
 \hline
\end{tabular}
\label{tab:params}
\end{table}
\begin{figure}
    \centering
    \includegraphics[width=\linewidth]{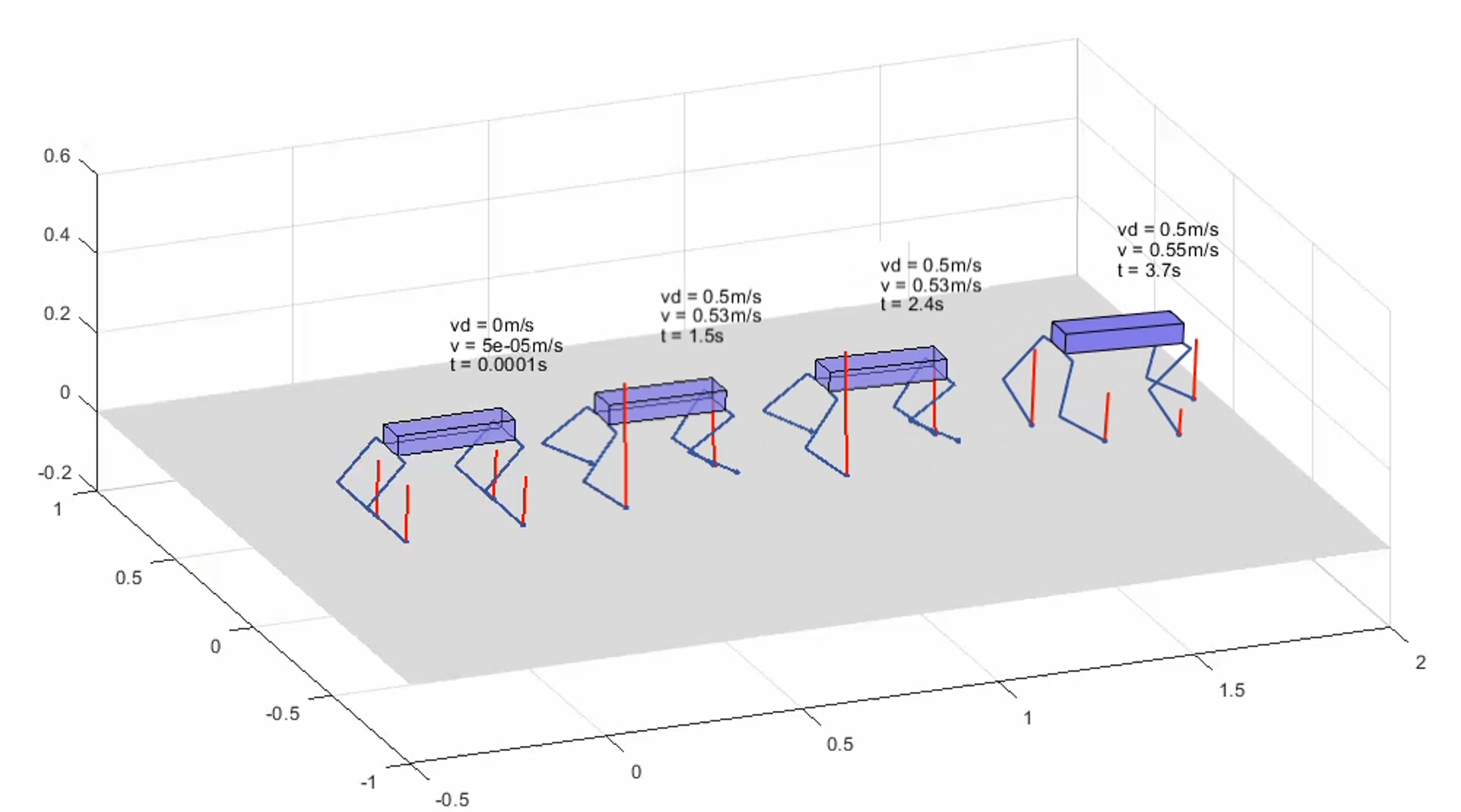}
    \caption{Simulation of a robot walking in a straight line with a trot gait using GRFs output from the MPC. The red line shows the GRFs of feet in contact during each phase of the trot gait}
    \label{fig:trot_walk}
\end{figure}
Figure \ref{fig:trot_walk} shows the robot walking in a straight line along the x-axis with the GRFs output from MPC. The z-direction foot forces while executing a trot gait are shown in Figure \ref{fig:trot_z}. The GRFs generated by the MPC controller match fairly with an ideal sequence. Here the ideal sequence of GRFs in the z-direction is $mg/n$ where $n$ is the number of feet in contact (computed using newtons laws). Further, the robot is able to execute a stable walk with a variety of gaits, such as crawl and bound, with the gait parameters listed in Table \ref{tab:gait}.

\begin{table}[ht]
\caption{Tuning Parameters}
\centering
\begin{tabular}{ c c c c } 
 \hline
 Parameter & Value \\
\hline
 $\bm{Q}_{p}$       & $\mathbb{I}_3\times 1e^6$ \\
 $\bm{Q}_{v}$       & $\mathbb{I}_3\times 1e^6$ \\
 $\bm{Q}_{\Theta}$  & $\mathbb{I}_3\times 1e^6$ \\
 $\bm{Q}_{\omega}$  & $\mathbb{I}_3\times 1e^6$ \\
 $\bm{K}$           & $\mathbb{I}_{12}\times 1e^1$ \\ 
 $\bm{K}_p^{sw}$         & 300 \\
 $\bm{K}_d^{sw}$         & 0.1 \\
 \hline
\end{tabular}
\label{tab:gains}
\end{table}

\begin{figure}
    \centering
    \includegraphics[width=\linewidth]{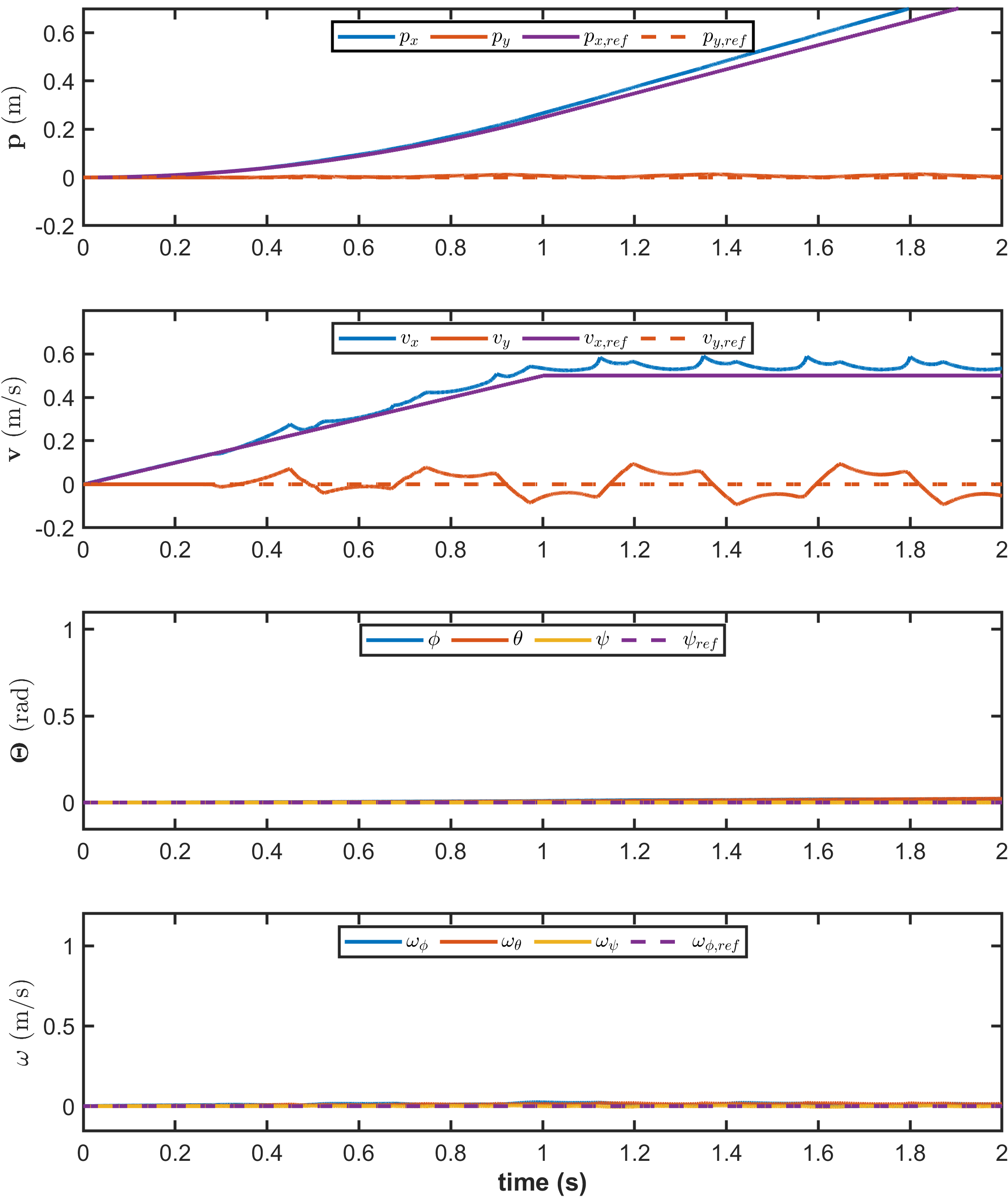}
    \caption{Robot states while executing a trot gait for robot walking in a straight line}
    \label{fig:trot_state}
\end{figure}

The state trajectories for a straight line trajectory while executing a trot gait with a command x-velocity of $v_{d,x} = 0.5 m/s$ with the desired acceleration of $\dot{v}_d = 0.5 m/s^2$ is shown in Figure \ref{fig:trot_state}. The MPC-based controller is able to track the desired velocity while the angular positions and velocities remain bounded. 

\begin{figure}
    \centering
    \includegraphics[width=\linewidth]{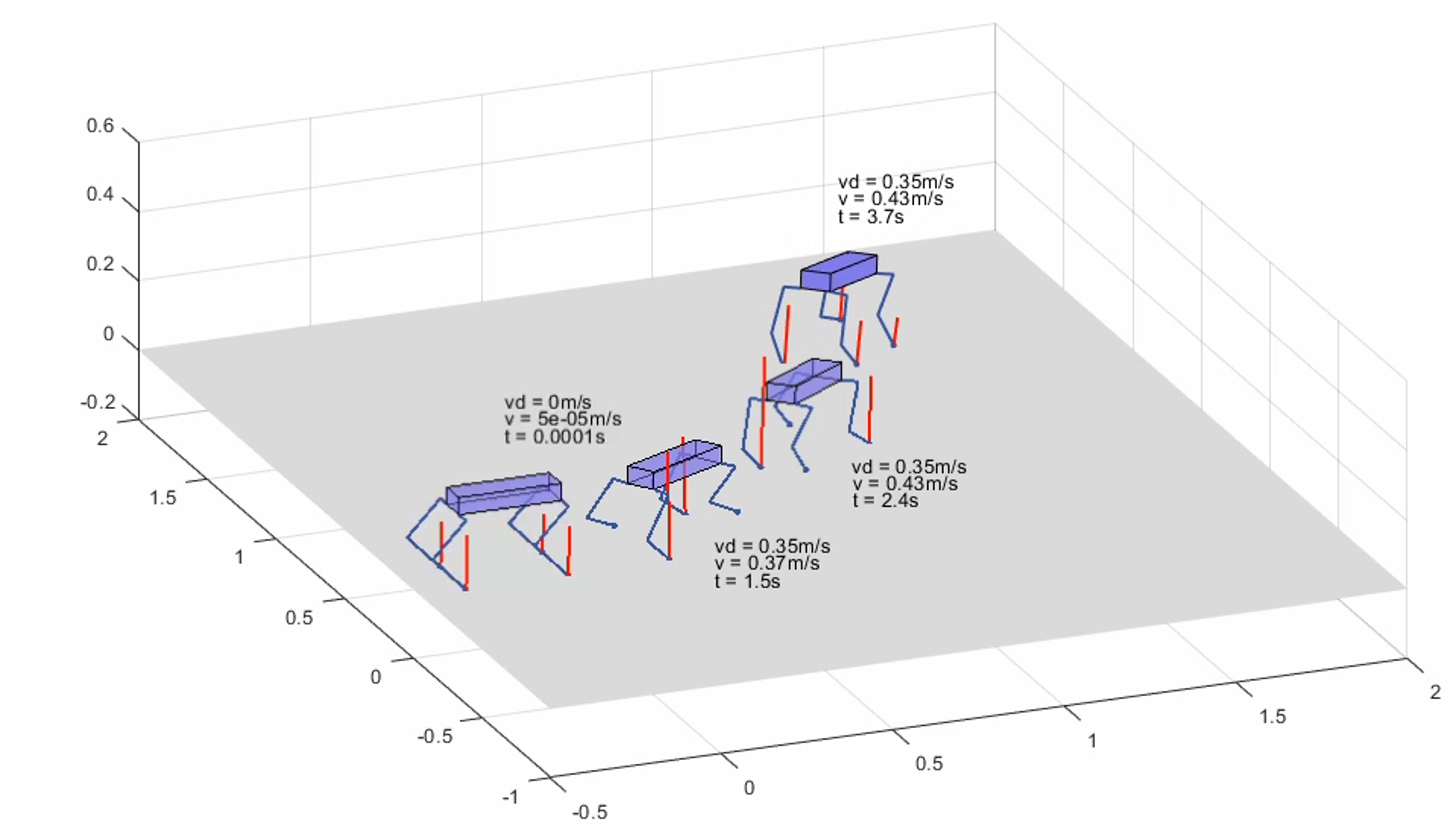}
    \caption{Simulation of a robot turning with $psi=pi/4$ with a trot gait using GRFs output from the MPC. The red line shows the GRFs of feet in contact during each phase of the trot gait}
    \label{fig:trot_turn}
\end{figure}

\begin{figure}
    \centering
    \includegraphics[width=\linewidth]{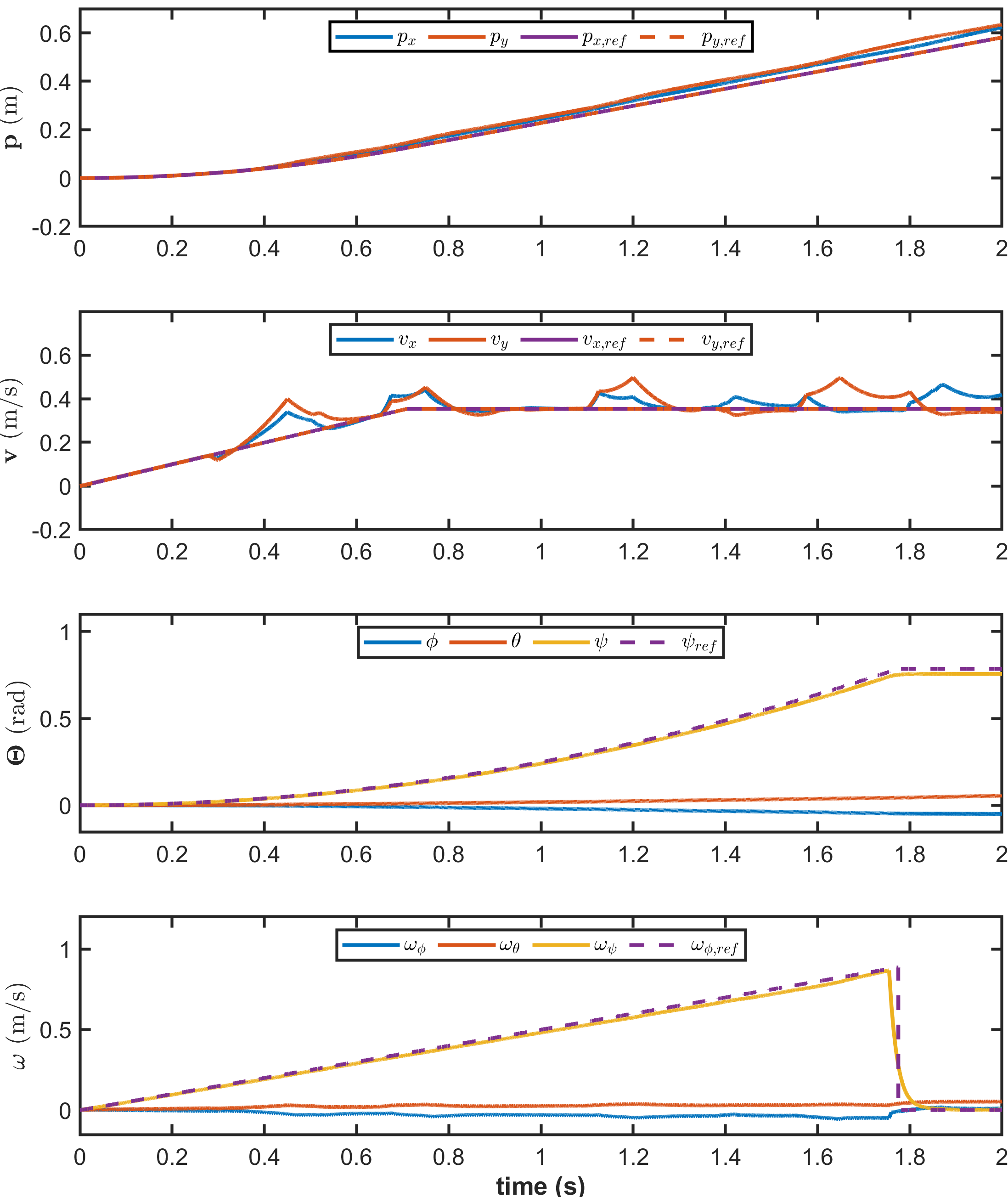}
    \caption{Robot states while executing a turn with $\psi=pi/4$ with trot gait}
    \label{fig:trot_state2}
\end{figure}

Figure \ref{fig:trot_turn} shows the robot executing a turn with yaw angle $\psi = pi/4$ with the desired velocity of $\bm{v}_d = 0.35 m/s$. The corresponding robot states are shown in Figure \ref{fig:trot_state2}. The robot is able to change its heading angle $\psi$ until it reaches $\psi_{ref}=pi/4$ while maintaining a constant desired velocity $\bm{v}_d$. The yaw angular velocity $\omega_\psi$ goes to zero as soon as the yaw angle $\psi$ meets the desired value of $\psi_d = pi/4$.

\begin{figure}
    \centering
    \includegraphics[width=\linewidth]{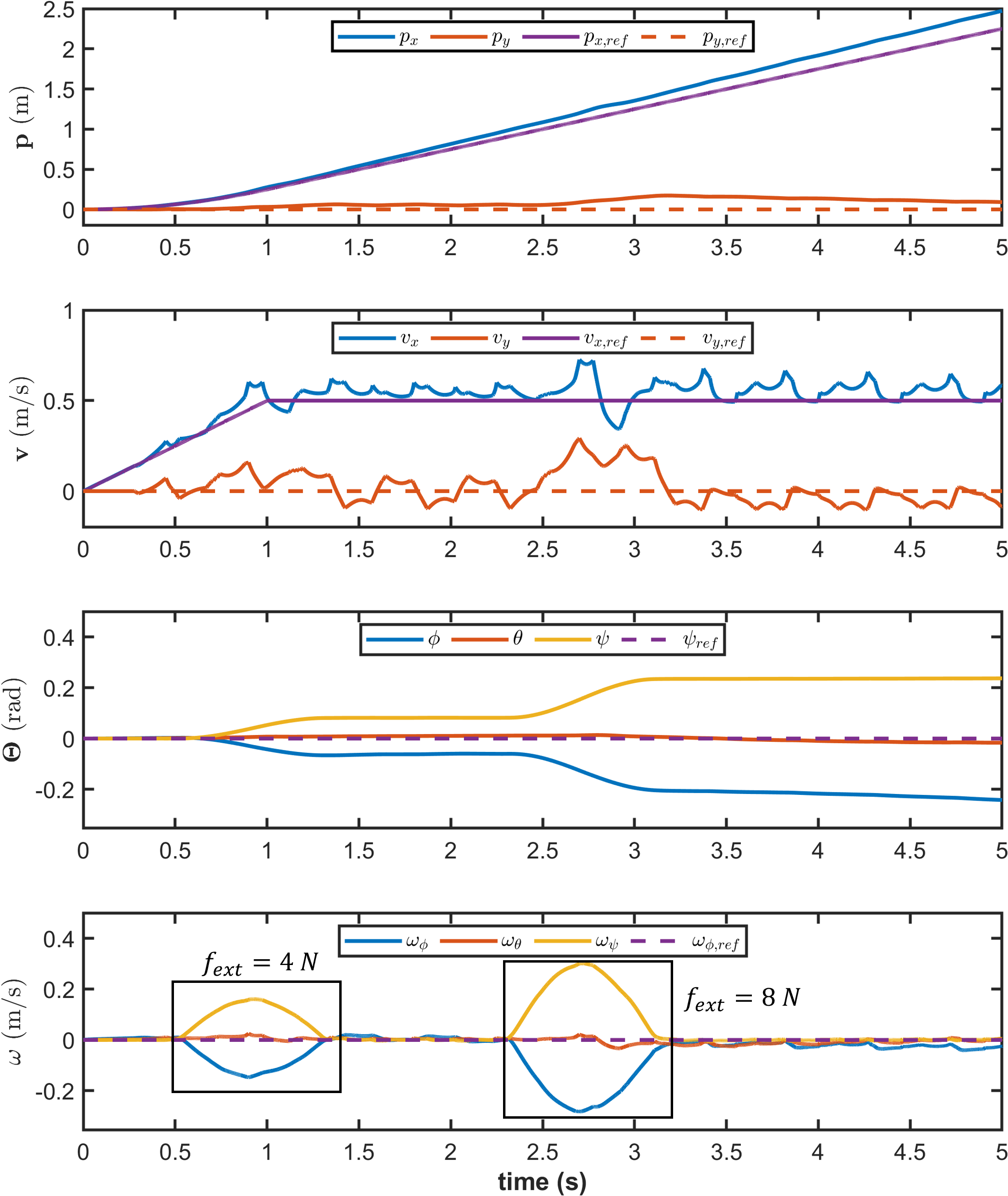}
    \caption{Robot states while executing a trot gait for robot walking in a straight line under unknown external disturbance forces represented as a Bezier polynomial. The external disturbance is executed at $t = 0.5$s under $f_{ext} = 4$N and $t = 2.3$s under $f_{ext} = 8$N.}
    \label{fig:states_trot_disturbance}
\end{figure}

Lastly, the performance of the robot under an unknown external disturbance $f_{ext} = 4$ N applied at the midpoint of the robot at $t= 0.5$s and $f_{ext} = 8$ N at $t = 2.3$s represented as a Bezier polynomial as shown in Figure \ref{fig:states_trot_disturbance} while executing a trot gait with a commanded x-velocity of $v_{d,x} = 0.5 m/s$ with the desired acceleration of $\dot{v}_d = 0.5 m/s^2$. The MPC-based controller is perturbed from its reference states under external forces and does not converge back to its initial state. This is due to the deviation $\phi$ from the small angle assumption in Section \ref{subsection: LTV Dynamics}. However, simulating for a longer duration ($t>15$ s) leads to error accumulation in the angular positions $\bm{\Theta}$ and $\bm{w}$. This could be attributed to the fact that the LTV formulation relies on small angle assumptions in $\phi$ and $\theta$ and does not account for the effect of angular motion on the rigid body. Long-term performance can be improved by incorporating error correction schemes and using a more accurate model. 

\section{CONCLUSIONS}
This work presents an LTV MPC for quadruped locomotion. Through a hierarchical control scheme, it is shown that given a gait planner and a simplified model (SRB), the optimal control formulation can be rewritten as a QP using $qpOASES$ through the $CasADi$ interface. This proposed LTV MPC is capable of tracking desired reference trajectories for different gaits such as trot and crawl. The proposed MPC can reach up to 1 m/s top speed with an acceleration of 0.5 m/s$^2$ executing a trot gait. Additionally, it is shown that under unknown external disturbances, the quadruped is able to stabilize and track the desired reference trajectories. Future works will include exporting the MPC in $CasADi$ to C code such that it can be integrated into the hardware and implementing motion planning algorithms. 

\addtolength{\textheight}{-12cm}   % This command serves to balance the column lengths
                                  % on the last page of the document manually. It shortens
                                  % the textheight of the last page by a suitable amount.
                                  % This command does not take effect until the next page
                                  % so it should come on the page before the last. Make
                                  % sure that you do not shorten the textheight too much.

%%%%%%%%%%%%%%%%%%%%%%%%%%%%%%%%%%%%%%%%%%%%%%%%%%%%%%%%%%%%%%%%%%%%%%%%%%%%%%%%

\section*{ACKNOWLEDGMENT}

The authors would like to thank Dr. Umesh Vaidya and DIRA lab for helping us formulate the quadruped locomotion problem and Dr. Vahidi for his insights into MPC design. Sriram contributed to formulating the LTV MPC, linearization methods, and implementing the LTV MPC using $CasADi$. Andrew contributed to trajectory generation, finite state machine, and simulation setup. Both authors contributed equally to the literature review, project planning, hierarchical control, experiment methods, and result analysis. 

%%%%%%%%%%%%%%%%%%%%%%%%%%%%%%%%%%%%%%%%%%%%%%%%%%%%%%%%%%%%%%%%%%%%%%%%%%%%%%%%

%Bibliography
\bibliographystyle{unsrt}  
\bibliography{references}  

\end{document}